\documentclass[journal]{IEEEtran}
\usepackage{graphicx}
\usepackage{subfigure}
\usepackage{mathrsfs}
\usepackage{amsmath}
\usepackage{multirow}
\usepackage{siunitx}
\usepackage{float}
\usepackage{colortbl}
\usepackage{xcolor}
\usepackage{soul}
\usepackage{bm}
\usepackage{cite}
\usepackage{balance}
\usepackage{stfloats}
\usepackage{algorithm}
\usepackage{algorithmicx}
\usepackage{algpseudocode}
\usepackage{amsmath}
\usepackage{epsfig}
\usepackage{hyperref}

\hypersetup{
	colorlinks=true,
	linkcolor=cyan,
	urlcolor=red,
	citecolor=cyan,
}

\definecolor{mygray}{gray}{.9}

\begin{document}

\title{Physics-Guided Generative Adversarial Networks for Sea Subsurface Temperature Prediction}

\author{Yuxin~Meng,~\emph{Student Member, IEEE,}
        Eric~Rigall,~\emph{Student Member, IEEE,}
        Xueen~Chen,~\emph{Member, IEEE,}
        Feng~Gao,~\emph{Member, IEEE,}
        Junyu~Dong,~\emph{Member, IEEE,}
				Sheng~Chen,~\emph{Fellow, IEEE}
\thanks{This work was supported in part by the National Key Research and Development Program of China (Grant 2018AAA0100602) and the National Key Scientific Instrument and Equipment Development Projects of China (Grant 41927805). \emph{(Corresponding authors: Junyu Dong, Feng Gao)}.} %
\thanks{The datasets and our code for this work are available in \emph{https://github.com/mengyuxin520/PGGAN.}} %
\thanks{Y. Meng, E. Rigall and F. Gao are with Department of Computer Science and Technology, Ocean University of China, Qingdao 266100, China (E-mails: mengyuxin520@126.com, rigall1214397@163.com, gaofeng@ouc.edu.cn).} %
\thanks{X. Chen is with College of Oceanic and Atmospheric Sciences, Ocean University of China, Qingdao 266100, China (E-mail: xchen@ouc.edu.cn).} %
\thanks{J. Dong is with Institute of Advanced Oceanography, Ocean University of China, Qingdao 266100, China (E-mail: dongjunyu@ouc.edu.cn).} %
\thanks{S.~Chen is with School of Electronics and Computer Science, University of Southampton, Southampton SO17 1BJ, UK (E-mail: sqc@ecs.soton.ac.uk).} %
\vspace*{-5mm}
}

\maketitle

\begin{abstract}
\textcolor{blue}{This work has been accepted by IEEE TNNLS for publication.}
Sea subsurface temperature, an essential component of aquatic wildlife, underwater dynamics and heat transfer with the sea surface, is affected by global warming in climate change. Existing research is commonly based on either physics-based numerical models or data based models. Physical modeling and machine learning are traditionally considered as two unrelated fields for the sea subsurface temperature prediction task, with very different scientific paradigms (physics-driven and data-driven). However, we believe both methods are complementary to each other. Physical modeling methods can offer the potential for extrapolation beyond observational conditions, while data-driven methods are flexible in adapting to data and are capable of detecting unexpected patterns. The combination of both approaches is very attractive and offers potential performance improvement. In this paper, we propose a novel framework based on generative adversarial network (GAN) combined with numerical model to predict sea subsurface temperature. First, a GAN-based model is used to learn the simplified physics between the surface temperature and the target subsurface temperature in numerical model. Then, observation data are used to calibrate the GAN-based model parameters to obtain better prediction. We evaluate the proposed framework by predicting daily sea subsurface temperature in the South China sea. Extensive experiments demonstrate the effectiveness of the proposed framework compared to existing state-of-the-art methods.
\end{abstract}


\IEEEpeerreviewmaketitle

\section{Introduction}\label{S1}

\IEEEPARstart{S}{ea} subsurface is the part of ocean below the sea surface. Its temperature plays an important role in ocean science research \cite{chen18_grsl}. Sea subsurface temperature is important information for understanding the global ocean ecosystem and earth climate system. The study of the spatial and temporal distribution of sea temperature and its variation law is not only a critical issue in marine geography, but also of considerable significance to fishery, navigation, and underwater acoustics. Diverse sources of external factors, such as radiation and diurnal wind, affect the sea subsurface temperature, and the prediction of the sea subsurface information is very challenging \cite{Mcphaden98}. Existing studies on sea subsurface temperature rely on numerical modeling and observational data \cite{tandeo09_grsl,meng20_grsl,wu12_grsl,hosoda12_grsl}.

Numerical modeling is a widely used technique to tackle complex ocean problems by data simulation, based on the equations of ocean physical laws. Currently, Princeton Ocean Model (POM) \cite{Umlauf03_jmr}, HYbrid Coordinate Ocean Model (HYCOM) \cite{Chassignet07_jms}, and Finite-Volume Coastal Ocean Model (FVCOM) \cite{Chen06} are commonly used in oceanography. POM is a classic traditional ocean model with clear structure, concise model specifications, and thorough model physical interpretation. The flexible vertical hierarchical structure of HYCOM makes it more suitable for the significant expansion of the stratification effect. FVCOM model includes momentum equation, continuity equation, thermo-salt conservation equation and state equation. The numerical solution of FVCOM adopts the finite volume method (FVM), which has the advantages of accurate and fast calculation and good fitting of coastline boundary and seabed topography based on the unstructured mesh. This is because FVM can better guarantee the conservation of each physical quantity not only in each unit but also in the whole calculation area. All these numerical models are constructed based on our knowledge of ocean physics, and they are often applied to simulate ocean dynamics and predict sea subsurface temperature. However, their prediction accuracy can hardly be guaranteed, since there exist a large range of environmental factors that affect marine environments.

In order to improve the prediction accuracy of the numerical models, assimilation methods are commonly used. Traditional assimilation methods can improve the model prediction performance by fusing new observational data in the dynamic running process of a numerical model. Smedstad and O’Brien \cite{smedstad} summarized the data assimilation methods developed before 1991 and classified them into polynomial interpolation methods, optimal interpolation methods, and variational analysis methods. Anderson \emph{et al.} \cite{anderson} also surveyed the data assimilation methods in Physical Oceanography. Although the prediction accuracy of the traditional assimilation methods is much higher than that of the numerical models, there are ample rooms that these methods can be further improved.

In contrast to the physics-based numerical models,  data-driven models, such as neural networks, rely purely on observational data to learn the underlying data distribution. However, it is unclear how these models produce specific decisions, and interpreting these data-driven models physically are very difficult. Since these methods only rely on training data, their generalization ability on unseen data is often limited, whereas most physics-based models do not utilize training data and therefore may perform well on unseen data, provided that the physical laws employed to build these models accurately represent the underlying data distribution. Nevertheless, the physical rules are often incomplete, and these numerical models need to be improved and supplemented.

A fundamental principle in data modeling is to incorporate available \emph{a priori} information regarding the underlying data generating mechanism into the modeling process. Data-physics hybrid models capable of incorporating prior knowledge typically outperforms  data-driven modeling \cite{HongChen2009,Chen_etal2011}. Motivated by this fundamental principle for data modeling, in this paper, we focus on developing a physics-guided framework for training neural network to predict sea subsurface temperature, which combines numerical modeling and observational data modeling. We demonstrate that this data-physics hybrid modeling approach can not only take advantage of our prior knowledge of ocean physical laws but also improve the overall prediction accuracy.

In recent years, deep learning in computer vision \cite{Cheng18_tgrs,Mou17_tgrs,Zhang18_tgrs} and natural language processing \cite{Li16_nips,Sarikaya14_taslp,Korpusik19_taslp} has achieved breakthrough progress. Its underlying motivation is to simulate the human brain neural connection structures \cite{liu_hang_17_tgrs,sun19_grsl,henry18_grsl,ogut19_tgrs}. When handling high-dimensional data, high-level features are  extracted through multiple layers progressively to identify the concepts relevant to human \cite{Chen16_tip,Wang20_tip,Perera19_tip}. Deep learning models can be roughly divided into two categories: discriminant models and generative models \cite{wang_liu_19_tgrs,wang_yuan_19_tgrs}. Discriminant models are trained to distinguish the correct output among possible output choices \cite{Yang19_tgrs,Sun20_tgrs}. On the other hand, generative models are trained to obtain better understandings of the data samples. Specifically, a generative model learns a distribution from the input samples, and then generates similar samples based on this distribution to enhance the model. Goodfellow {\it et al.} \cite{goodfellow14_nips} proposed the generative adversarial network (GAN), which uses adversarial training to train a generative network and a discriminative network jointly. The generative network captures the potential distribution of the real data, while the discriminative network is commonly a binary classifier which judges whether the input samples are real or not. Many GAN-based models have been proposed to solve the problem of high-quality image generation. Isola {\it et al.} \cite{Isola17_cvpr} proposed Pix2pix for image translation. In Pix2pix, a pair of image datasets from different domains are fed into the model during training, and an image can be transformed from one domain to the other. Zhu {\it et al.} \cite{Zhu17_cvpr} proposed CycleGAN to learn mappings between an input image and an output image when paired training data is unavailable. A cycle consistency loss is introduced to achieve this goal.

\begin{figure}[tp!]
\begin{center}
 \includegraphics[width=0.98\columnwidth,angle=0]{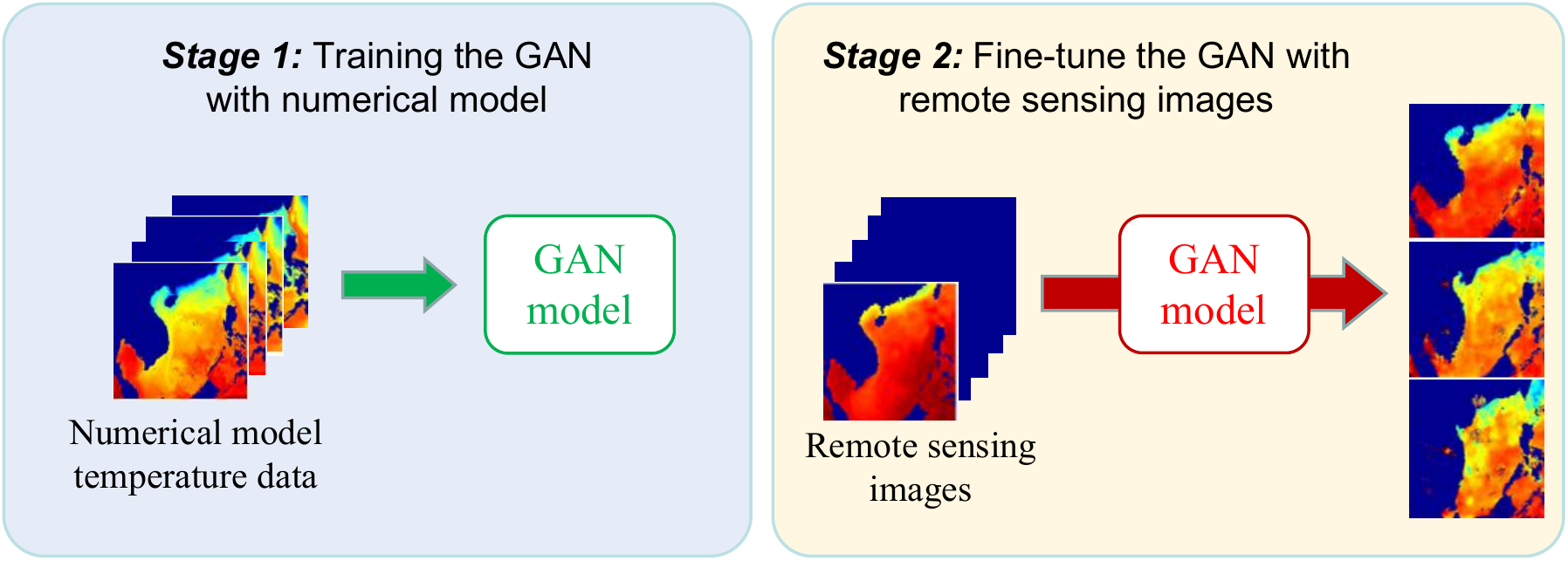}
\end{center}
\vspace*{-4mm}
\caption{Two-stage sea subsurface temperature prediction framework. In the first stage, generative adversarial training is performed on the model with the data from the physics-based numerical model. In the second stage, the model is fine-tuned with observational data.}
\label{fig_1} 
\vspace*{-4mm}
\end{figure}

The deep neural network has strong predictive power but it does not follow the laws of physics. By contrast, a numerical model simulates the ocean dynamics, based on knowledge of ocean physics. Karpatne {\it et al.} \cite{Karpatne17_arxiv} blended the numerical model with multi-layer perceptron to correct lake temperature. In this work, the authors applied all the variables related to the lake temperature and the output of the numerical model for the lake temperature as the inputs to the neural network. If the numerical model  accurately simulates the motion of the lake temperature, the output of their model is generated by the numerical model; otherwise the result is generated by the neural network. This approach basically chooses the result from either the physics-based numerical model or the neural network trained by observation data. Ideally, we would like to design a prediction method by combining both the physics-based numerical model and the data-driven model. Jia {\it et al.} \cite{jia_PGRNN} combined a recurrent neural network (RNN) model with the numerical model to predict the lake temperature. Their model was trained over the numerical model data and then fine-tuned on the limited observation data. However, their model was applied for each depth separately, and the data from the same depth is used to predict the lake temperature of the same depth. In addition, they only predicted the temperature value at one subsurface point, not over an entire area.  We also note that most existing studies concentrate on the sea surface prediction, while there is a paucity of contributions on the daily sea subsurface temperature prediction. This will be further discussed in the related work section.

\begin{figure*}[!bp]
\vspace*{-5mm}
\begin{center}
\includegraphics[width=0.9\linewidth,angle=0]{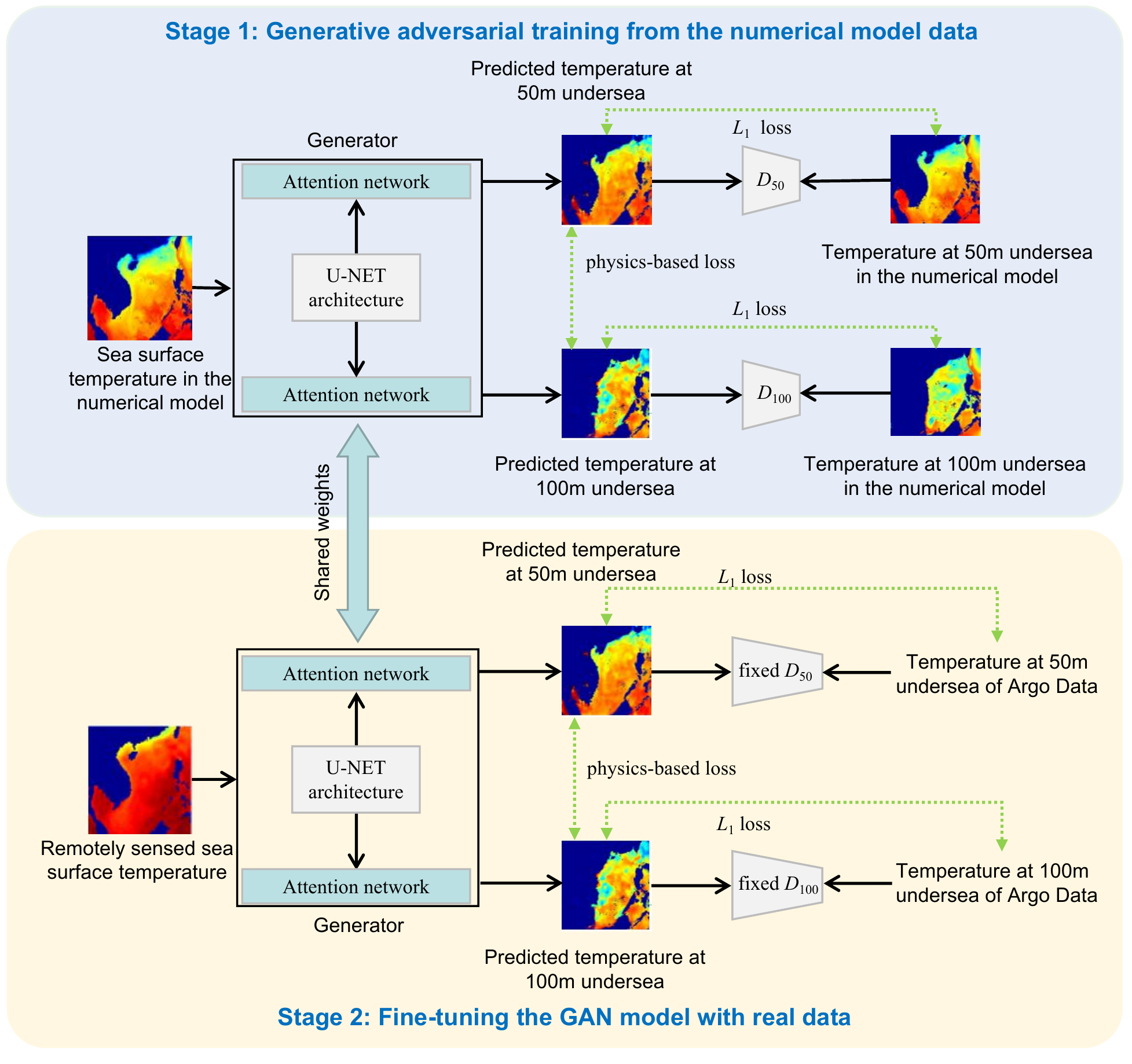}
\end{center}
\vspace*{-7mm}
\caption{The proposed GAN-based sea subsurface temperature prediction framework. \emph{Stage} 1: The generator learns the mapping from the sea surface temperature to the target depth temperature in the numerical model. The generator is composed of two components: one single shared network and several task-specific attention networks. The shared network learns the mapping from sea surface temperature and random noise to the numerical model data. The task-specific attention networks capture the mapping between the sea surface temperature and the sea subsurface temperature. \emph{Stage} 2: Fine-tuning the GAN model with observational data. The weights of the generator are shared with Stage 1, and the weights of the discriminator are fixed.}
\label{fig_framework} 
\vspace*{-2mm}
\end{figure*}

In this paper, to tackle the above-mentioned limitations in the existing sea temperature analysis literature, we propose a new framework to predict the sea subsurface temperature by combining the physics-based numerical model with deep neural networks. In our method, we apply the physics-based numerical model to train the neural network model in the first phase, and then observational data is used to calibrate the model parameters in the second phase. More specifically, we design two neural networks in the proposed framework, as illustrated in Fig.~\ref{fig_1}. The first network learns the simplified physics laws from the numerical model. The weights of this first network are shared by the second network. This effectively encodes the knowledge of ocean physics into this second network model, and its weights are then fine-tuned by observational data. It can be seen that the merits of both physics-based numerical modeling and observational data modeling approaches are combined and, consequently, the prediction accuracy is further enhanced. The main contributions of this paper are summarized as follows.
\begin{itemize}
\item A novel GAN-based framework is proposed which predicts the daily sea subsurface temperature by learning the relationship between sea surface temperature and subsurface temperature.
\item We explore the use of GAN combined with the physics-based numerical model for building a hybrid prediction model incorporating more effectively the known ocean physics with the observational data information.
\item We propose a physics-based loss with a mask as prior knowledge. The mask filters out land locations and this loss automatically encodes the knowledge of ocean physics into the modeling process, leading to prediction performance improvement.
\end{itemize}

The rest of the paper is organized as follows. Section~\ref{S2} presents the background of GAN models and sea temperature prediction. Section~\ref{S3} details the proposed framework for sea subsurface temperature prediction. The experimental results are reported in Section~\ref{S4}. We draw concluding remarks and discuss the future work in Section~\ref{S5}.

\section{Related Works}\label{S2}

\subsection{Generative Adversarial Networks}\label{S2.1}

Inspired by the binary zero-sum game, Goodfellow {\it et al.} \cite{goodfellow14_nips} proposed GAN in which two neural networks contest each other in a game. More specifically, GAN is composed of two networks: a generative network $G$ and a discriminative network $D$. The generator $G$ iteratively learns the distribution of the real input samples, and it generates samples following the learnt distribution. The generated fake samples are then fed into the discriminator $D$, and $D$ is trained to judge whether the input samples are real or fake.

In the training process, the generator $G$ learns the input data distribution. During this learning process, fake samples can be identified by the discriminator $D$ from the real data distribution. In such an adversarial learning, the generator $G$ tries to `fool' the discriminator $D$ by producing samples as similar as possible to the real samples. With this mutual competitive reinforcement, the performances of both $G$ and $D$ are jointly enhanced.

Conditional generative adversarial network (CGAN) \cite{mirza14} is an extension of GAN in which a conditional setting is applied. In CGAN, both the generator $G$ and discriminator $D$ are conditioned on class labels. As a result, the model can learn mappings from inputs to outputs by feeding it with contextual information.

Yang {\it et al.} \cite{PI_SDE} solved the stochastic differential equations by encoding the known physical laws into the GAN. L{\"u}tjens {\it et al.} \cite{PI_CFV} used GAN to learn the latent features of the numerical model data in order to generate more realistic coastal floor data. Zheng {\it et al.} \cite{PI_SI} reconstructed the image based on its known pixels by employing a GAN model. These works used the GAN model to learn the latent features from the numerical model. Then they applied the pre-trained GAN model to do the corresponding tasks. In other words, these works used the GAN models to replace part or the entire numerical model. The works \cite{PI_SDE,PI_CFV,PI_SI} highlight the potential application of the GAN model in physical-relevant tasks. However, the difference of these works with our hybrid physics-data based GAN is huge. Not only we pre-train the GAN with the physics-based numerical model but also we adopt the observational data to calibrate the pre-trained GAN model. In other words, our GAN model not only learns the physical laws from the numerical model but also adapts itself using observational data.

\subsection{Sea Subsurface Temperature Prediction}\label{S2.2}

Temperature is an important factor in marine hydrology and climate change \cite{liu17_tgrs}. Existing studies based on satellite remote sensing data mainly focus on sea surface temperature and assessment. Yang {\it et al.} \cite{yang18_grsl} considers the task of sea temperature prediction as a sequence prediction problem and builds an end-to-end trainable long short-term memory (LSTM) neural network model. Then, the temporal and spatial features are combined to predict sea temperature. Wei {\it et al.} \cite{wei20_grsl} used Ice Analysis (OSTIA) data to train a neural network for South China Sea temperature prediction. Deep learning-based methods have also been utilized to predict the sea surface temperature in Bohai Sea and Indian Ocean \cite{zhang17_grsl,patil17_od,patil18_jaot}.

\begin{figure*}[!tp]
\vspace*{-2mm}
\begin{center}
\includegraphics[width=0.8\linewidth,angle=0]{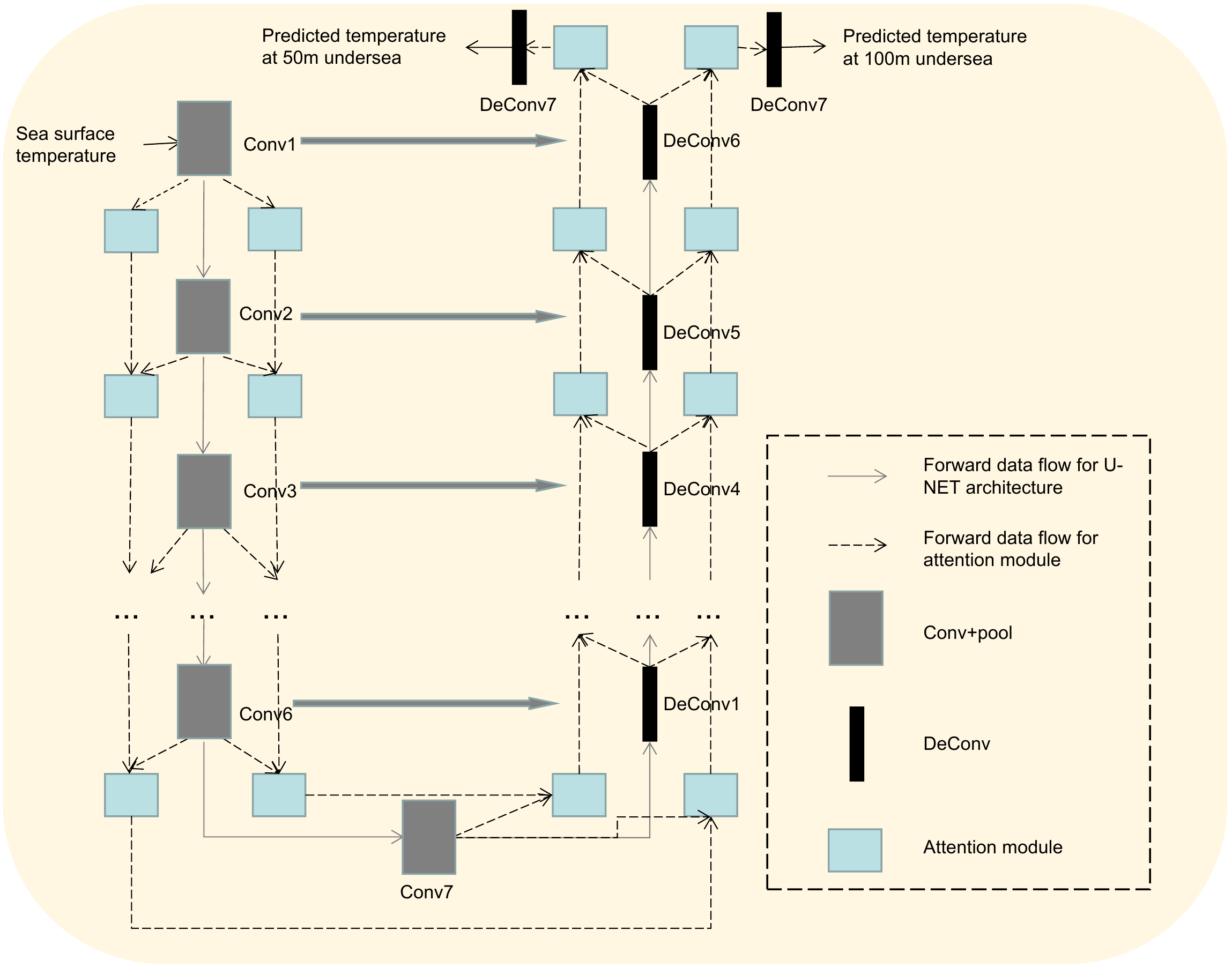}
\end{center}
\vspace*{-6mm}
\caption{Illustration of the generator architecture. The generator comprises of the U-NET architecture and the two sets of attention module. The attention module is connected with the output of the convolution block and the attention module from the last layer, which is passed one by one.}
\label{fig_generator} 
\vspace*{-4mm}
\end{figure*}

The above mentioned studies mainly focus on temperature prediction of the sea surface. However, the sea subsurface temperature prediction research is scarce.  Han {\it et al.}  \cite{Conv_SST} applied the convolutional neural network (CNN) to predict the subsurface temperature from a sets of the remote sensing data. Lu {\it et al.} \cite{Cnn_SST} adopted the pre-clustered neural network method to estimate the subsurface temperature and the results are better than those obtained without clustering. Wu {\it et al.} \cite{Snn_SST} used the self-organizing map neural network to predict the subsurface temperature anomaly in the North Atlantic. These methods can reliably predict the monthly subsurface temperature using neural network owing to the fact that sufficient monthly observational data of the subsurface temperature are available for training neural network models. However, due to the very limited daily observation data, the prediction of the daily subsurface temperature cannot be carried out efficiently and accurately only using neural networks. Zhang {\it et al.} \cite{zhang19_grsl} used monthly Argo data to predict the sea subsurface temperature but no physics-based numerical model was utilized in this monthly sea subsurface temperature prediction model. These works indicate the lack of research on daily subsurface temperature prediction. In this paper, we combine deep neural networks and  a physics-based numerical model into a unified framework, which is capable of predicting the daily sea subsurface temperature.

\section{Proposed Framework}\label{S3}

The proposed framework, depicted in Fig.~\ref{fig_framework}, composes of two stages: 1)~generative adversarial pre-training on numerical model data; and 2)~fine-tuning of the GAN model with observational data. In the first stage, the generator learns the mapping from the sea surface temperature to the target depth temperature using numerical model data. This effectively encodes the knowledge of ocean physics into the neural network model. In the second stage, real-world observation data are used to fine-tune the weights of the neural network model. This enables the model to learn the real-data distribution and to compensate for physics knowledge missing in the numeral model. We now detail these two stages.

\subsection{Stage~1:~Generative Adversarial Training on Numerical Model Data}\label{S3.1}

Numerical models play an important role in understanding the ocean's influence on global climate. They simulate the ocean properties and circulation based on the equations of ocean physics laws. Since numerical models approximate the physical correlations among different depths of the ocean, we use a GAN model in the proposed framework to acquire these relationships from the data generated by a numerical model.

Without loss of generality, we consider predicting the subsurface temperatures at 50m, 100m and 150m underwater simultaneously. The prediction tasks of different depth temperatures can be achieved jointly by multi-task learning. In order to obtain good performance for each task, attention modules are used to enable both the task-shared and task-specific feature learning in an end-to-end manner \cite{liu_multitask}. The generator architecture is depicted in Fig.~\ref{fig_generator}, which is comprised of multiple sets of attention modules and the U-NET architecture. Each set of attention module can learn the features for individual tasks. Specifically, each attention module learns a soft attention mask, which is dependent on the features in the shared network. The features in the shared network and the soft attention masks can be trained jointly to optimize the generalization of the features across different tasks.

\begin{figure}[tp!]
\begin{center}
\includegraphics[width=0.98\columnwidth,angle=0]{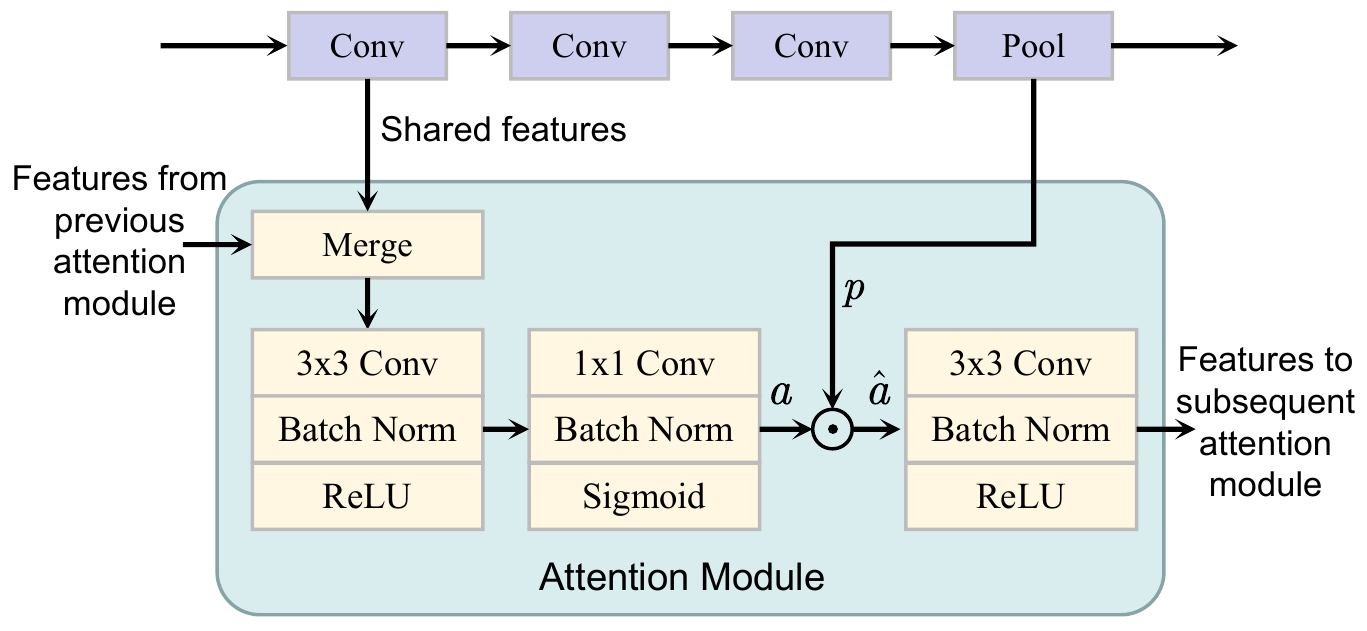}
\end{center}
\vspace*{-4mm}
\caption{Illustration of the attention module.}
\label{fig_attention_module} 
\vspace*{-4mm}
\end{figure}

As shown in Fig.~\ref{fig_attention_module}, the shared features after pooling are denoted as $p$, and the learnt attention mask in the layer for task $i$ is denoted as $a_i$. The task-specific features $\hat{a}_i$ are computed by element-wise multiplication of the attention mask with the shared features as $\hat{a}_i\! =\! a_i\! \odot\! p$, where $\odot$ denotes element-wise multiplication operator. The attention module has strong capabilities of emphasizing non-trivial features and weakening unimportant ones. Moreover, as the seawater temperature generally decreases with the increase of depth. we exploit this fact and set it as prior knowledge. If the seawater temperature in a lower layer is estimated higher than the one in an upper layer, the model is penalized. Hence we apply this physics-based loss to guide the fitting ability of the model between different depths\footnote{In some high latitude oceanic regions, seawater temperature at 50\,m can actually be higher than sea surface temperature. In this case, the physics-based loss should not be applied to this first underwater layer.}.

As mentioned in Subsection~\ref{S2.1}, a GAN model is composed of two networks: the generative network $G$ and the discriminative network $D$. In our model, the generator contains two parts: one single shared network, and three task-specific attention networks. The shared network uses a conditional GAN model which learns a mapping from the observed image $x$ and random noise $z$ to real image $y$. The network objective is defined as follows:
\begin{align}\label{eq1}
L_{S_1}(G, D) =& E_{x,y}\log D(x,y)  \nonumber \\
& + E_{x,z}\log(1-D(x, G(x,z))) ,
\end{align}
where $E_{x,y}$ denotes the expectation operator with respect to $x$ and $y$, $D(x,y)$ distinguishes whether $x$ and $y$ are the true paired data, and $G(x,z)$ learns the mapping from the data $x$ and random vector $z$ to the real data $y$. As can be observed in Eq.\,(\ref{eq1}), in the shared network, the generator $G$ tries to minimize the objective while $D$ tries to maximize it. In the generator model, the input noise $z$ and conditional information $x$ jointly constitute the joint hidden layer representation in order to model the same distribution with domain $y$. To further improve the prediction performance, we mix the conditional GAN objective with a $L_1$ distance which is defined as
\begin{equation}\label{eq2}
  L_{L_1}(G) = E_{x,y,z}\lVert G(x,z)-y \rVert_1.
\end{equation}

Besides the shared network, we build three task-specific attention networks, $G_{0\rightarrow50}$, $G_{0\rightarrow100}$ and $G_{0\rightarrow150}$, to capture the mappings between the sea surface temperature and the undersea temperatures at 50m, 100m and 150m, respectively. Correspondingly, the discriminative network can be decomposed into three sub-discriminative networks, namely, $D=\{D_{50},D_{100},D_{150}\}$.

In our implementation, the sea surface temperature $x_0$ is obtained from the HYCOM data \cite{Chassignet07_jms}. Besides $x_0$, we generate three masks, $M_{0}$, $ M_{50}$ and $ M_{100}$.  Starting from $M_0$, its value at a given location is set to 1 if the sea surface temperature is available from the numerical model at this location, and the value is set to 0 if the temperature is not exploitable, e.g., the location is on the land. This mask can filter out noise regions, such as the land. We further set the margin to 0.1. If the temperature of the deep layer is 0.1 degree higher than that of the shallow layer, the model is penalized. Specifically, we define an objective function $L_{0\sim50}(G)$ as follows:
\begin{align}\label{eq3}
& L_{0\sim50}(G) = \nonumber \\
& \hspace*{3mm} E_{x,z}\lVert\max\{(G_{0\rightarrow50}(x_{0},z)
\odot M_{0}\! -\! x_{0}\odot M_{0}), 0.1\}\rVert_1.\!
\end{align}
The purpose of the mask $M_0$ can be seen clearly from the objective function $L_{0\sim50}(G)$. Only when the temperature at the 0\,m depth is exploitable, i.e., this location is not on land, the comparison between the temperature at the -50\,m depth and the temperature at the 0\,m depth is meaningful. Similarly, we have the mask $M_{50}$, whose value at a location is set to 0 if the numerical model data indicates that this 50\,m depth location is on the land; otherwise the temperature at this location is exploitable and we set $M_{50}=1$. Likewise, we can calculate $M_{100}$. Hence we can define the objectives $L_{50\sim100}(G)$ and $L_{100\sim150}(G)$ respectively as:
\begin{align}
L_{50\sim100}(G) =& E_{x,z}\lVert\max\{(G_{0\rightarrow100}(x_{0},z)\odot M_{50}-  \nonumber \\
& \hspace*{7mm} G_{0\rightarrow50}(x_{0},z)\odot M_{50}), 0.1\}\rVert_1 , \label{eq4} \\
L_{100\sim150}(G) =& E_{x,z} \lVert\max\{( G_{0\rightarrow 150}(x_{0},z) \odot M_{100}- \nonumber \\
& \hspace*{7mm} G_{0\rightarrow100}(x_{0},z)\odot M_{100}), 0.1\}\rVert_1. \label{eq5}
\end{align}
Based on the above three objective functions, we propose the physics-based loss by using the three masks as prior knowledge, which leads to an improved prediction performance. Hence the physics-based loss in Stage 1 is defined as:
\begin{align}\label{eq6}
L_{P_1}(G) =& L_{0\sim50}(G) + L_{50\sim100}(G) + L_{100\sim150}(G).
\end{align}

It can be seen that this physics-based loss applies `shallower sea temperature' as masks (SL masks). Specifically, when comparing the temperature difference of a deeper layer and the shallower layer, the mask is referencing the shallower-layer sea temperature. It is natural to ask whether we can adopt `deeper sea temperature' as masks (DP masks). That is, when comparing the temperature difference of a deeper layer and the shallower layer, the mask is referencing the deeper layer sea temperature. Adopting DP masks in the proposed physics-based loss corresponds to replacing $M_0$ in the loss (\ref{eq3}) with $M_{50}$ and replacing $M_{50}$ in the loss (\ref{eq4}) with $M_{100}$ as well as calculating the mask $M_{150}$ and using it to replace $M_{100}$ in the loss (\ref{eq5}). This DP mask approach however is less effective than the SL mask approach. This is because the land area at the 0\,m depth is smaller than that at the -50\,m depth, and the land area at the -50\,m depth is smaller than that at the -100\,m depth, and so on. Therefore, the exploitable values of $M_0$ ($M_0=1$) or the size of $M_0$ is much larger than that of $M_{50}$, the size of $M_{50}$ is much larger than that of $M_{100}$, and the size of $M_{100}$ is much large than that of $M_{150}$. Hence, adopting the SL mask approach enables the model to exploit larger sea subsurface area. In the ablation study of Subsection~IV-C, we will demonstrate that better performance can be obtained by adopting the SL mask approach than the DP mask approach.

By employing a physics-based loss, the generator can learn the mapping from the sea surface temperature to the temperature at 50m, 100m and 150m undersea from the numerical model simultaneously. According to the prior knowledge, the sea surface temperature should be higher than the one at 50m undersea, which should be higher than the temperature at 100m underwater and so on. If there is some irregular data, the penalty term will be added in the training process.

According to the above discussion, the full objective function in the first stage of generative adversarial training using the physics-based numerical model data is expressed as
\begin{equation}\label{eq7}
L(G, D)=L_{S_1}(G, D) + L_{L_1}(G) + L_{P_1}(G).
\end{equation}

\begin{algorithm}[tp!]
\caption{Stage I training procedure}
\label{ALG1}
\begin{algorithmic}[1]
\Require
HYCOM model training data $x$, $y$, random noise vector $z$, sea temperature masks $M_{0}$, $M_{50}$ and $M_{100}$,  initial learning rate $l_1$, learning rate decaying factor $\eta$, numbers of critic iterations $n_1$, $n_2$
\Require
Initial generator parameters $\theta_{g}$, initial discriminator parameters $\theta_d=\{\theta_{d}^i\}_{i\in\{50,100,150\}}$
\Ensure
{Generator\,$G$\,and\,discriminator\,${D\! =\!\{D_{i}\}_{i\in\{50,100,150\}}}$}
\While {not converged}
  \State Set learning rate to $l=l_1$;
  \For{$t=0,\cdots,n_1$}
    \State Sample image pair $\{x_{0}^{i}\}_{i=1}^{N}$ and $\{y_{50}^{i}\}_{i=1}^{N}$, $\{x_{0}^{i}\}_{i=1}^{N}$ and $\{y_{100}^{i}\}_{i=1}^{N}$, $\{x_{0}^{i}\}_{i=1}^{N}$ and $\{y_{150}^{i}\}_{i=1}^{N}$;
    \State Update $D$ by gradient descent based on cost (\ref{eq1});
    \State Update $G$ by gradient descent based on cost (\ref{eq7});
  \EndFor
  \For{$t=n_1+1,\cdots,n_1+n_2$}
    \State Sample image pair $\{x_{0}^{i}\}_{i=1}^{N}$ and $\{y_{50}^{i}\}_{i=1}^{N}$, $\{x_{0}^{i}\}_{i=1}^{N}$ and $\{y_{100}^{i}\}_{i=1}^{N}$, $\{x_{0}^{i}\}_{i=1}^{N}$ and $\{y_{150}^{i}\}_{i=1}^{N}$;
    \State Update $D$ by gradient descent based on cost (\ref{eq1});
    \State Update $G$ by gradient descent based on cost (\ref{eq7});
    \State Update learning rate $l=l_1-\eta(t-n_1)$;
  \EndFor
\EndWhile
\end{algorithmic}
\end{algorithm}

Algorithm~\ref{ALG1} implements the first stage of the training process in our proposed method. The weights of the discriminators and the generator are updated based on the costs (\ref{eq1}) and (\ref{eq7}) separately. In our implementation, the first $ n_1\! =\! 100$ epochs maintain a constant learning rate of $l_1\! =\!0.0002$,  followed by another $n_2\! =\! 100$ epochs with a linearly decaying learning rate whose decaying factor $\eta$ satisfies $0 < \eta <\frac{l_1}{n_2}$. This setting is the same as the original Pix2Pix method \cite{Isola17_cvpr}.

\subsection{Stage~2:~Fine-tuning GAN Model with Observation Data}\label{S3.2}

Since numerical models rely heavily on simplified physics law, their results sometimes exhibit discrepancies from the observed data. Therefore, we utilized remotely sensing data, Argo data \cite{argo_data},  to correct numerical data errors.

As illustrated in Fig.~\ref{fig_framework}, AVHRR Sea Surface Temperature (SST) data \cite{sst_data} is fed as the input of the model, while Argo data is employed as the real data. The generator shares the weights with the model from the first stage, while the weights of the discriminators are fixed. The generator in the second stage is composed of one single shared network and two task-specific attention networks. The objective function of the shared network is as follows:
\begin{equation}\label{eq8}
L_{S_2}(G) = E_{x,z}\log(1-D(x, G(x,z))) ,
\end{equation}
where the discriminator $D$ does not update its weights, and only the generator updates its parameters through backpropagation. In this stage, the real data is Argo data. As Argo data contains the temperature information at single location, we cannot use Argo data to train the discriminator. Instead, we have to fix the discriminator in order to predict the temperatures on the entire area, not at a point location like Argo data. Since daily Argo data are point data, to predict values from the point to the entire plane, the discriminator pretrained in the first stage is used to measure the differences between the generated samples and the real data according to
\begin{equation}\label{eq9}
L_{dot}(G)=E_{x,z} \lVert G(x_{i,j},z)-\textrm{Argo}_{i,j} \rVert_1,
\end{equation}
where the index pairs $i$ and $j$ denote the locations of the temperature values from Argo data. Since daily Argo data only contain one temperature value, we employ $L_1$ distance to measure the temperature error between Argo data and the corresponding generated sample in (\ref{eq9}). By doing this, the adjustment from point to plane can be achieved.

\begin{algorithm}[bp!]
\caption{Stage II training procedure}
\label{ALG2}
\begin{algorithmic}[1]
\Require
Remote sensing satellite training data $x$, Argo training data $\text{Argo}_{50}$, $\text{Argo}_{100}$ and $\text{Argo}_{150}$, random noise vector $z$, sea temperature masks $M_{50}$ and $M_{100}$, initial learning rate $l_1$, learning rate decaying factor $\eta$, numbers of critic iterations $n_{1}$, $n_{2}$
\Require
Generator parameters $\theta_{g}$ and discriminator parameters $\theta_d=\{\theta_{d}^i\}_{i\in\{50,100,150\}}$
\Ensure
Generator $G$
\While {not converged}
  \State Set learning rate to $l=l_1$;
  \For{$t=0,\cdots,n_1$}
    \State Sample image pair $\{x_{0}^{i}\}_{i=1}^{N}$ and $\{\text{Argo}_{50}^{i}\}_{i=1}^{N}$, $\{x_{0}^{i}\}_{i=1}^{N}$ and $\{\textbf{Argo}_{100}^{i}\}_{i=1}^{N}$, $\{x_{0}^{i}\}_{i=1}^{N}$ and $\{\text{Argo}_{150}^{i}\}_{i=1}^{N}$;
    \State Update $G$ by gradient descent based on cost (\ref{eq11});
  \EndFor
  \For{$t=n_1+1,\cdots,n_1+n_{2}$}
    \State Sample image pair $\{x_{0}^{i}\}_{i=1}^{N}$ and $\{\text{Argo}_{50}^{i}\}_{i=1}^{N}$, $\{x_{0}^{i}\}_{i=1}^{N}$ and $\{\text{Argo}_{100}^{i}\}_{i=1}^{N}$, $\{x_{0}^{i}\}_{i=1}^{N}$ and $\{\text{Argo}_{150}^{i}\}_{i=1}^{N}$;
    \State Update $G$ by gradient  descent based on cost (\ref{eq11});
    \State Update learning rate $l=l_1-\eta(t-n_1)$;
  \EndFor
\EndWhile
\end{algorithmic}
\end{algorithm}

It should be noted that in the second stage, two task-specific attention networks are employed. Due to the imprecision of AVHRR SST data, the temperature difference between the sea surface and 50m undersea is not taken into account. In the experiment section this will be fully explained. Therefore, the physics-based loss in Stage 2 is defined as:
\begin{align}\label{eq10}
L_{P_2}(G) = L_{50\sim100}(G) + L_{100\sim150}(G),
\end{align}
where $L_{50\sim100}(G)$ and $L_{100\sim150}(G)$ use the same configurations as the corresponding objective functions in the first stage.

The full objective function employed in the second stage is therefore given by:
\begin{equation}\label{eq11}
L(G) = L_{S_2}(G) + L_{dot}(G) + L_{P_2}(G).
\end{equation}

Algorithm~\ref{ALG2} implements the second stage of the training process in our method.

\section{Experiments}\label{S4}

\subsection{Study Area and Data}\label{S4.1}

The study was conducted on South China sea, a marginal sea in the western Pacific Ocean, located in the south of Mainland China. The sea has an area of about 3.5 million square kilometers, with an average depth of 1,212 meters and a maximum depth of 5,559 meters. A typical study area of (\ang{3.99}N$\sim$\ang{24.78}N, \ang{98.4}E$\sim$\ang{124.4}E) was selected.

The numerical model data, satellite remote sensing data and Argo data from May 2007 to November 2017 were used for training. The remote sensing data from January 2004 to April 2007 were employed as the test input data. The Argo data from January 2004 to April 2007 were used as the true values for the comparison with the predictions, i.e., in the testing, the predicted results are compared with the Argo data.

The numerical model data used in our experiments is HYCOM from \cite{hycom_data}. The HYCOM data format is NetCDF and its spatial resolution is  1/\ang{12}$\times$1/\ang{12}. The data is configured with 32 layers in the vertical direction.

The National Oceanic and Atmospheric Administration (NOAA) optimum interpolation SST (OISST) data from \cite{sst_data} is used in this paper. The spatial resolution of the SST data is \ang{0.25} $\times$ \ang{0.25}, and daily mean data is employed in our study.

The Argo data employed in our study is collected from \cite{argo_data}. The Argo data is composed of the data collected from different buoys placed at different locations in the South China sea. The daily Argo data is sea subsurface temperature data acquired at only one point in the whole sea area. As the Argo data are point data, we randomly choose one point from the predicted temperature results at the target locations to compare with the true value of the Argo data at the same locations.

More specifically, the sea temperatures of the numerical model data at the depths of 0\,m, -50\,m, -100\,m and -150\,m are used for the first training stage. In the second training stage, we train the model over the satellite remote sensing data and the Argo data at the depths of -50\,m, -100\,m, -150\,m. The input data for the first stage is constructed in the format: [3856,\,128,\,128,\,1], where the first number is the size of the training dataset, the next two numbers are the height and the width of the input data, respectively, and the last number represents the grey-scale map with one color channel. Similarly, the format for the input data is [2020,\,128,\,128,\,1] in the second training stage, after removing the bad quality Argo data. In the test stage, the formats of the input data and the output data are [180,\,128,\,128,\,1] and [540,\,128,\,128,\,1], respectively, where the output data includes equal numbers of data samples for the sea subsurface temperature at 50\,m, 100\,m and 150\,m.

\subsection{Baseline Models and Evaluation Metrics}\label{S4.2}

To the best of our knowledge, this paper is the first to predict daily sea subsurface temperature by using methods other than numerical models. Due to the sparsity of the observational sea subsurface temperature data for training, it is not feasible to predict the temperature of a whole ocean area by solely relying on neural network.  Therefore, in our experimental evaluation, we combine the neural network methods with the numerical model and the traditional data assimilation approach to perform study. Since there are only a few Argo devices in the entire China South Sea, daily sea temperature can only be obtained in a small set of data points in the entire region. Thus, our method adopts numerical model data to do predictions first due to limited observational data, as the numerical model can simulate ocean dynamics and obtain sea temperature in the entire region. Then we use the set of observational data to fine-tune the model. In the experiments, when we need to compare with other state-of-the-art methods, we also train the model with numerical model data using those methods and then observational data are applied for fine-tuning. The data generated by the numerical model assimilation method are obtained from \cite{hycom_data}. This HYCOM assimilation data has a spatial resolution of 1/\ang{12}$\times$1/\ang{12}, a temporal resolution of 1 day, a vertical resolution from the sea surface to 5000 meters undersea. It is much closer to the observational data compared than the HYCOM model data. We compare these data with our method in the following experiment part. 

Furthermore, the following neural network methods are selected as the baselines to compare with our model: Pix2pix \cite{Isola17_cvpr}, CycleGAN \cite{Zhu17_cvpr}, and PGNN \cite{Karpatne17_arxiv}. For Pix2pix and CycleGAN, we use the publicly available source codes provided by the authors, with the same default parameters. Specifically, for Pix2pix, $\lambda=100$ and $70\times 70$ PatchGAN are employed as mentioned in \cite{Isola17_cvpr}. For CycleGAN, an Adam solver \cite{kingma15_iclr} is employed with a learning rate of 0.0002. For PGNN, its output comes from either the neural network or the numerical model. However, having an output solely relying on a pure neural network is unsuitable for daily sea subsurface prediction over the whole sea area. Therefore, we cannot directly compare the PGNN with our method. Since PGNN uses a physics-based loss to guide the training of its neural network, in our experiments, we compare the physics-based loss obtained by PGNN with the physics-based loss obtained by our method. Additionally, we also compare our method with the methods of \cite{jia_PGRNN} and \cite{zhang19_grsl}.

The two evaluation criteria, the root mean square error (RMSE) and the coefficient of determination ($R^2$) \cite{R2}, are used to assess the performance of the compared methods.

\subsection{Experiment Design and Ablation Study}\label{S4.3}

All our experiments are implemented on an NVIDIA GeForce 2080Ti GPU. Training iterations and learning rates are the same for the both phases of our approach. We train our model for $n_1\! +\! n_2\! =\! 200$ epochs. The first $n_1\! =\! 100$ epochs maintain a constant learning rate of $0.0002$, followed by another $n_2\! =\! 100$ epochs with a linearly decaying learning rate. The main network of the generator adopts a U-NET architecture \cite{UNET}, and each convolution is followed by an attention module. The discriminator applies the same six-layer convolutional network as in pix2pix \cite{Isola17_cvpr}. We construct the data as $128\times 128$ squared-shape heatmaps. Due to the inconsistency of Argo data underwater position, one-dimensional interpolation method was applied to obtain the data of 50 meters, 100 meters and 150 meters underwater. We use the Z-score standardization method to preprocess the data.

\begin{table}[tp!]
\caption{Study on the multi-task learning}
\label{table_multi_task} 
\vspace*{-4mm}
\begin{center}
\begin{tabular}{|c||c|c|c|}
\hline
\multirow{2}{*}{Model}  & \multicolumn{3}{c|}{RMSE ($^{\circ}$C)} \\
\cline{2-4} & ~~~~50m~~~~ & ~~~~100m~~~~ & ~~~~150m~~~~ \\ \hline\hline
Model without TANs & 0.9532 & 1.3265 & 1.2475 \\
Model with TANs~~~ & 0.9435 & 1.3067 & 1.2439 \\ \hline \hline
\multirow{2}{*}{Model}  & \multicolumn{3}{c|}{$R^{2}$} \\
\cline{2-4} & 50m & 100m & 150m \\ \hline   \hline
Model without TANs & 0.5431 & 0.3129 & 0.5410   \\
Model with TANs~~~~ & 0.5437 & 0.3374 & 0.5514   \\ \hline
\end{tabular}
\end{center}
\vspace*{-6mm}
\end{table}

We perform an extensive ablation study to demonstrate the effectiveness of the multi-task learning and physics-based loss. The influence of different margin values in physics-based loss is also studied. Moreover, the temperature difference between sea surface and 50m undersea is analyzed in detail.

\subsubsection*{Effectiveness of Multi-task Learning} 

Multi-task learning exploits the correlation among different tasks to promote each other, and consequently the performance of the whole model is enhanced. We add multiple task-specific attention networks (TANs) to learn the mappings from the sea surface temperature to 50m, 100m, and 150m undersea simultaneously. Table~\ref{table_multi_task} illustrates the RMSE and $R^2$ results on the usefulness of TANs. By using TANs, the RSME values improve 0.0097, 0.0198, and 0.0036, respectively, for predicting the sea subsurface temperatures 50m, 100m, and 150m undersea. Using TANs also improves the $R^2$ values. The results of Table~\ref{table_multi_task} therefore demonstrate that multi-task learning is effective to improve the prediction performance.

\subsubsection*{Effectiveness of the Mask}

When we employ the physics-based loss to guide the network training, the temperature between the upper and lower layers are compared by using a mask and the margin is set to 0.1 here. Here we compare several schemes: no use of mask (NO mask), deeper-layer sea temperature as mask (DP mask) and shallower-layer sea temperature as mask (SL mask).  Table~\ref{table_pbl} summarizes the RMSE and $R^2$ results obtained with these mask schemes. It can be seen that the method with the SL mask achieves the best RMSE and $R^2$ values. Therefore, we adopt the SL mask in the physics-based loss for our approach (see (\ref{eq3}) to (\ref{eq5}) and the discussions after (\ref{eq6})).

\begin{table}[hbp!]
\vspace*{-3mm}
\caption{Study on the mask in physics-based loss}
\label{table_pbl} 
\vspace*{-4mm}
\begin{center}
\begin{tabular}{|c||c|c|c|}
\hline
\multirow{2}{*}{Method}  & \multicolumn{3}{c|}{RMSE ($^{\circ}$C)}\\
\cline{2-4} & ~~~~50m~~~~ & ~~~~100m~~~~ & ~~~~150m~~~~ \\
\hline   \hline
~~~~ NO mask ~~~~    & 0.9480 & 1.3663 & 1.1977 \\
 DP mask     & 0.9647 & 1.3114 & 1.2048 \\
 SL mask     & 0.9333 & 1.2931 & 1.1969
\\ \hline \hline
\multirow{2}{*}{Method}  & \multicolumn{3}{c|}{$R^{2}$} \\
\cline{2-4} & 50m & 100m & 150m \\ \hline   \hline
 NO mask     & 0.5276 & 0.3020 & 0.5865   \\
 DP mask     & 0.5217 & 0.3277 & 0.5742   \\
 SL mask     & 0.5457 &0.3577 & 0.5885   \\ \hline
\end{tabular}
\end{center}
\vspace*{-3mm}
\end{table}

\subsubsection*{Analysis of the Margin} 

Likewise, in order to obtain better fitting ability, we add a margin in physics-based loss. First, we calculated the maximum temperature difference between the samples of two depths. Then the margin of the physics-based loss is scaled from 0 to Max. Table~\ref{table_margin} shows the prediction results of different margins. The best RMSE and $R^{2}$ values are obtained when the margin is set to 0.100. Therefore, in our approach we set the margin to 0.100 (see (\ref{eq3}) to (\ref{eq5})).

\begin{table}[tp!]
\vspace*{-1mm}
\caption{Study on different margins}
\label{table_margin} 
\vspace*{-4mm}
\begin{center}
\begin{tabular}{|c||c|c|c|}
\hline
\multirow{2}{*}{Margin}  & \multicolumn{3}{c|}{RMSE ($^{\circ}$C)} \\
\cline{2-4} & ~~~~ 50m ~~~~ & ~~~~ 100m ~~~~ & ~~~~ 150m ~~~~   \\ \hline   \hline
 zero      & 0.9976 & 1.3668 & 1.2345 \\
 0.001 & 0.9401 & 1.3077 & 1.2152 \\
 0.010 & 0.9731 & 1.3631 & 1.2418 \\
 0.100 & 0.9333 & 1.2931 & 1.1969 \\
 max   & 0.9403 & 1.3297 & 1.2063 \\ \hline \hline
\multirow{2}{*}{Margin}  & \multicolumn{3}{c|}{$R^{2}$} \\
\cline{2-4} & 50m & 100m & 150m   \\ \hline   \hline
 zero    & 0.4768 & 0.2795 & 0.5596   \\
 0.001 & 0.5397 & 0.3489 & 0.5597   \\
 0.010  & 0.5236 & 0.2813 & 0.5596   \\
 0.100   & 0.5457 & 0.3577 & 0.5885   \\
 max   & 0.5300 &0.3166 & 0.5831   \\ \hline
\end{tabular}
\end{center}
\vspace*{-5mm}
\end{table}

\subsubsection*{Analysis of the Physics-based Loss in Stage 2} 

In the second phase of the proposed method, we apply remote sensing data and Argo data to fine-tune the model. We estimate the contribution of $L_{0\rightarrow50}(G)$ in the physics-based loss in Table~\ref{table_loss_stage2}. It can be observed that the model without $L_{0\rightarrow50}(G)$ performs better. The reason is owing to the the imprecision of remote sensing AVHRR SST data, which degrades the performance of the model with $L_{0\rightarrow50}(G)$. Therefore, in our proposed method, we do not take $L_{0\rightarrow50}(G)$ into account in the physics-based loss in the second stage (see (\ref{eq10})).

\begin{table}[htp!]
\vspace*{-3mm}
\caption{Study on the physics-based Loss in Stage 2}
\label{table_loss_stage2} 
\vspace*{-4mm}
\begin{center}
\begin{tabular}{|c||c|c|c|}
\hline
\multirow{2}{*}{Method}  & \multicolumn{3}{c|}{RMSE($^{\circ}$C)} \\
\cline{2-4} & ~50m~ & ~100m~ & ~150m~ \\
\hline\hline
Model with $L_{0\rightarrow50}(G)$ in stage 2~~~ & 0.9465 & 1.3386 & 1.2031 \\
Model without $L_{0\rightarrow50}(G)$ in stage 2   & 0.9333 & 1.2931 & 1.1969 \\ \hline\hline
\multirow{2}{*}{Method}  & \multicolumn{3}{c|}{$R^{2}$} \\
\cline{2-4} & 50m & 100m & 150m \\ \hline   \hline
Model with $L_{0\rightarrow50}(G)$ in stage 2~~~   & 0.5218 & 0.3210 & 0.5826   \\
Model without $L_{0\rightarrow50}(G)$ in stage 2   & 0.5457 & 0.3577 & 0.5885   \\ \hline
\end{tabular}
\end{center}
\vspace*{-2mm}
\end{table}

\begin{table}[bp!]
\vspace*{-6mm}
\caption{Study on the network architecture}
\label{table_archi} 
\vspace*{-4mm}
\begin{center}
\begin{tabular}{|c||c|c|c|}
\hline
\multirow{2}{*}{Method}  & \multicolumn{3}{c|}{RMSE ($^{\circ}$C)}\\
\cline{2-4} & ~~~~50m~~~~ & ~~~~100m~~~~ & ~~~~150m~~~~ \\
\hline   \hline
 One attention module  & 0.9562 & 1.3286 & 1.2003 \\
 One discriminator     & 1.0075 & 1.2965 & 1.2010 \\
 Our method     & 0.9333 & 1.2931 & 1.1969
\\ \hline \hline
\multirow{2}{*}{Method}  & \multicolumn{3}{c|}{$R^{2}$} \\
\cline{2-4} & 50m & 100m & 150m \\ \hline   \hline
 One attention module     & 0.5429 & 0.2799 & 0.5741   \\
 One discriminator     & 0.4765 & 0.3511 & 0.5736   \\
 Our method     & 0.5457 &0.3577 & 0.5885   \\ \hline
\end{tabular}
\end{center}
\vspace*{-2mm}
\end{table}

\subsubsection*{Network Architecture Design} 

We use 3 attention modules and 3 discriminators for the three specific tasks, respectively. Considering the similarity in these tasks, the network architecture designs that exploit one attention module or one discriminator to learn different tasks are also experimented, and the results obtained are compared with our design in Table~\ref{table_archi}. The experimental results show that using more attention modules and discriminators can achieve better performance. Although our model performs better in this study than the model with single attention module and single discriminator, it has a higher computational complexity than the latter. In the case of predicting the subsurface temperatures at more than three depths, a single attention module with a single discriminator may become a better choice.

\begin{figure}[bp!]
\vspace*{-6mm}
\begin{center}
\includegraphics[width=0.88\columnwidth]{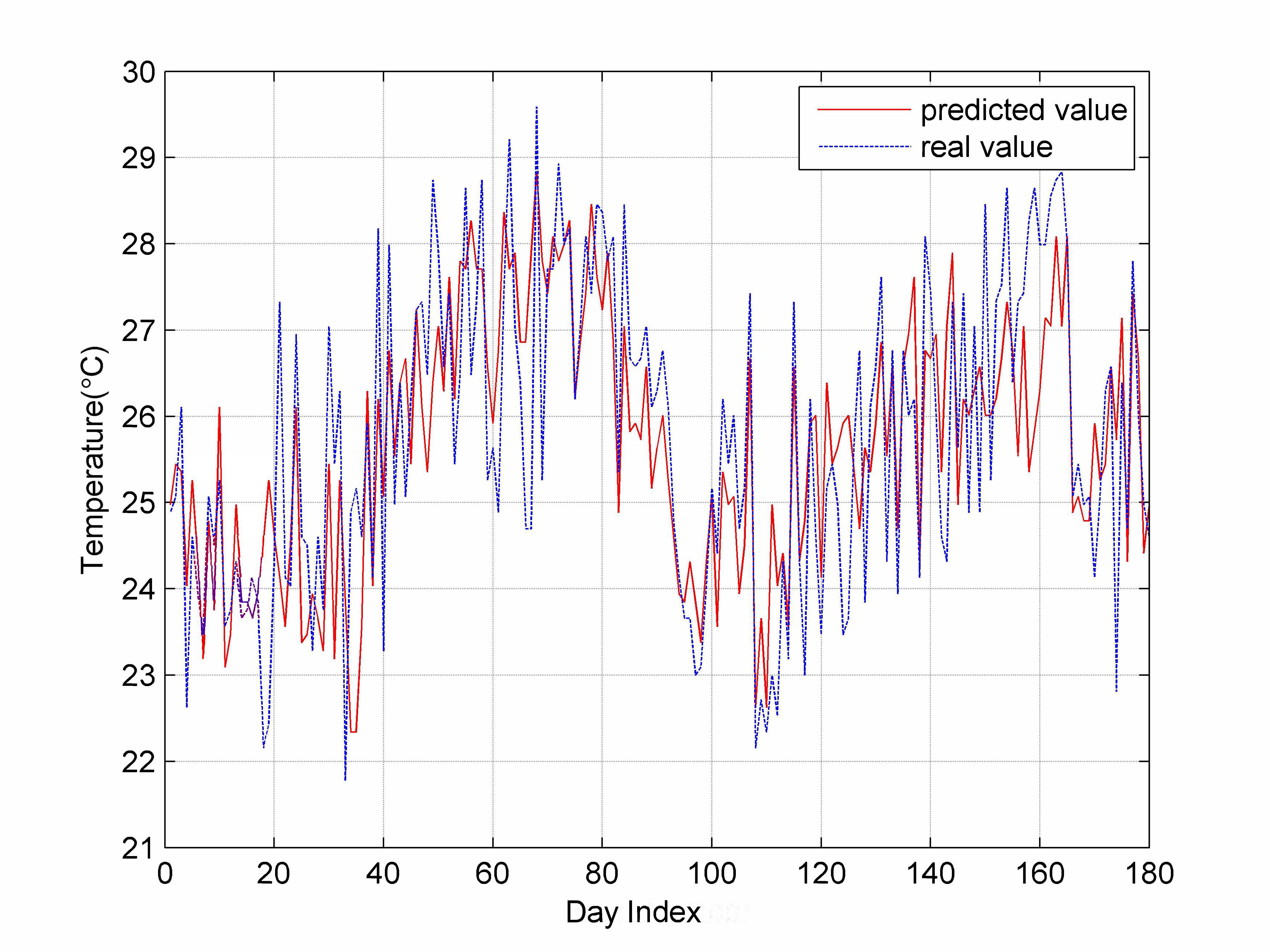}
\end{center}
\vspace*{-6mm}
\caption{Predicted temperature at 50m undersea versus Argo data.}
\label{fig_model_argo_50m} 
\vspace*{-2mm}
\end{figure}

\begin{figure}[bhp!]
\vspace*{-2mm}
\begin{center}
\includegraphics[width=0.88\columnwidth]{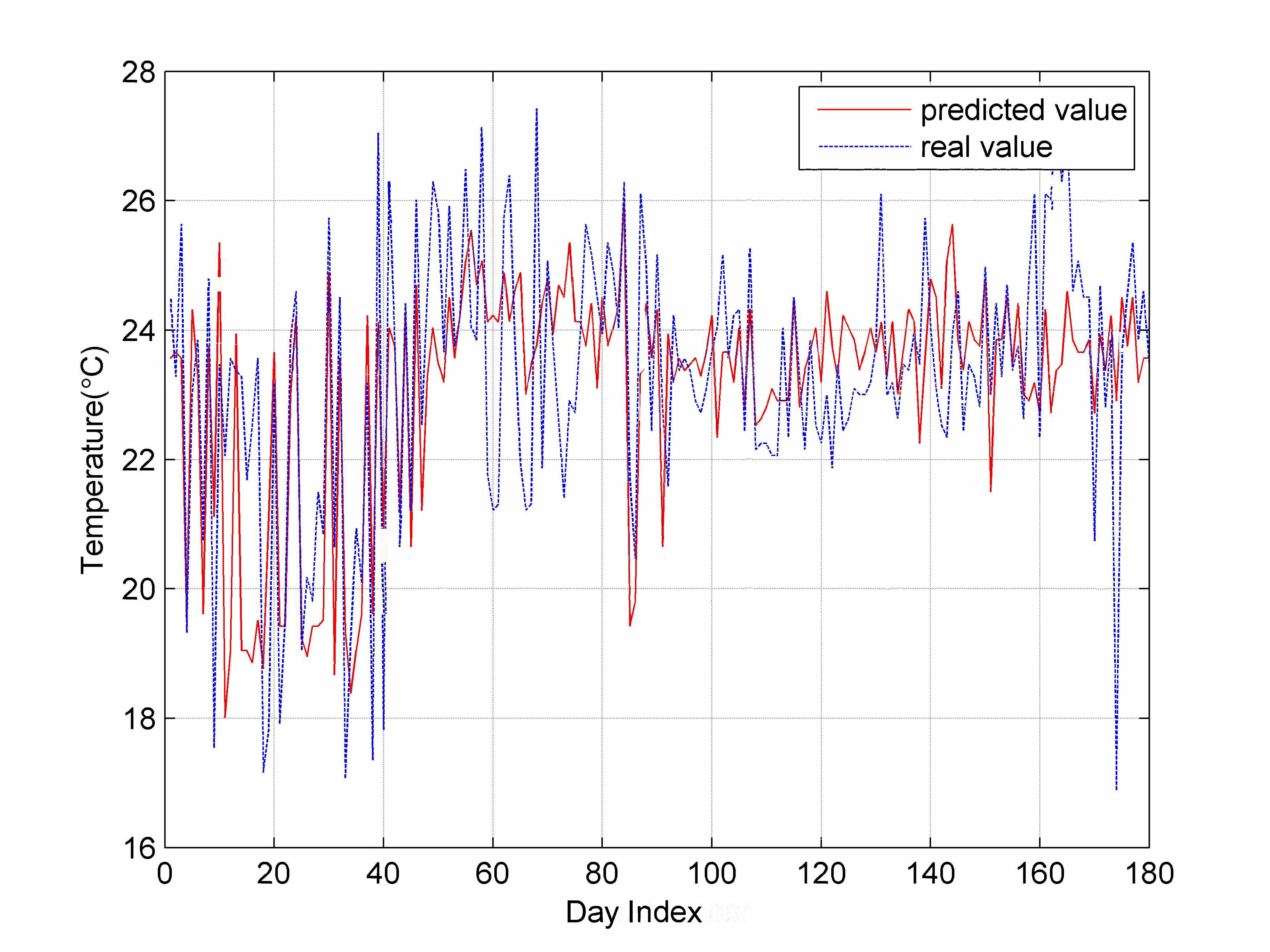}
\end{center}
\vspace*{-6mm}
\caption{Predicted temperature at 100m undersea versus Argo data.}
\label{fig_model_argo_100m} 
\vspace*{-2mm}
\end{figure}

\begin{figure}[bhp!]
\vspace*{-2mm}
\begin{center}
\includegraphics[width=0.88\columnwidth]{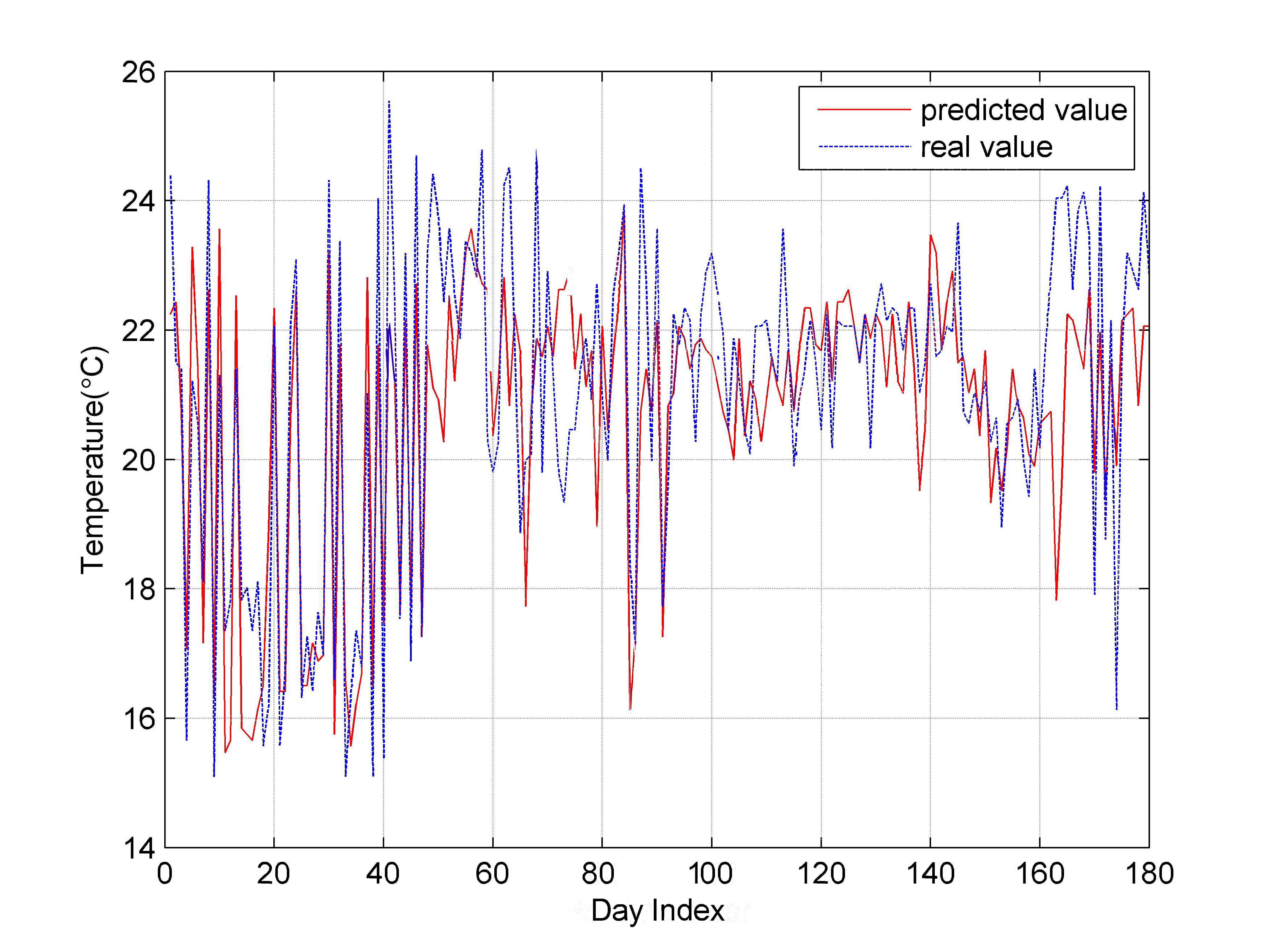}
\end{center}
\vspace*{-6mm}
\caption{Predicted temperature at 150m undersea versus Argo data.}
 \label{fig_model_argo_150m} 
\vspace*{-1mm}
\end{figure}

\begin{table*}[bp!]
\vspace*{-4mm}
\caption{Sea Subsurface Temperature Prediction Results (Average$\pm$STD) of Different Methods Averaged over 10 Random Runs}
\label{table_res_all} 
\vspace*{-4mm}
\begin{center}
\begin{tabular}{|c||c|c|c|c|c|c|}
\hline
\multirow{2}{*}{Model}  & \multicolumn{3}{c|}{RMSE($^{\circ}$C)} \\
\cline{2-4} & ~~~50m~~~ & ~~~100m~~~ & ~~~150m~~~  \\ \hline   \hline
Assimilation method    & 1.4520 & 1.8201 & 1.6774  \\
PGpix2pix  & 0.9528$\pm$0.0114 & 1.3301$\pm$0.024  & 1.2890$\pm$0.0415 \\
PGcycleGAN & 2.6155$\pm$0.0812 & 2.5345$\pm$0.042  & 2.7954$\pm$0.180 \\
PGNN       & 0.9482$\pm$0.0070 & 1.3691$\pm$0.0259 & 1.2837$\pm$0.039 \\
PGConvLSTM & 1.9213$\pm$0.223  & 1.6928$\pm$0.021  & 1.9974$\pm$0.127 \\
PGsim      & 1.1132$\pm$0.033  & 1.4659$\pm$0.083  & 1.3281$\pm$0.004 \\
Our method without PLoss & 0.9517$\pm$0.0082 & 1.3312$\pm$0.0251 & 1.2648$\pm$0.0313 \\
Our method with PLoss & \textbf{0.9402$\pm$0.0069} & \textbf{1.2894$\pm$0.0038} & \textbf{1.2330$\pm$0.0361} \\
\hline\hline
\multirow{2}{*}{Model} & \multicolumn{3}{c|}{$R^{2}$} \\
\cline{2-4} & 50m & 100m & 150m  \\ \hline   \hline
Assimilation method  & -0.4393  & -0.2661 & 0.1938   \\
PGpix2pix  & 0.5447$\pm$0.0119  & 0.3581$\pm$0.051   & 0.4992$\pm$0.043 \\
PGcycleGAN & -2.8694$\pm$0.44   & -1.9791$\pm$0.2798 & -1.0427$\pm$0.4027 \\
PGNN       & 0.5381$\pm$0.0112  & 0.2621$\pm$0.2089  & 0.2583$\pm$0.1958 \\
PGConvLSTM & -0.8927$\pm$0.2110 & -0.4445$\pm$0.3627 & -0.0655$\pm$0.1212 \\
PGsim      & 0.3555$\pm$0.0191  & 0.0179$\pm$0.2842  & 0.4514$\pm$0.042 \\
Our method without PLoss & 0.5512$\pm$0.0175 & 0.2934$\pm$0.045 & 0.5515$\pm$0.0212 \\
Our method with PLoss  & \textbf{0.5610$\pm$0.0153} & \textbf{0.3957$\pm$0.0392} & \textbf{0.5665$\pm$0.024} \\
\hline
\end{tabular}
\end{center}
\vspace*{-2mm}
\end{table*}

\begin{figure}[thp!]
\vspace*{-2mm}
\begin{center}
\includegraphics[width=0.77\columnwidth,height=0.77\columnwidth]{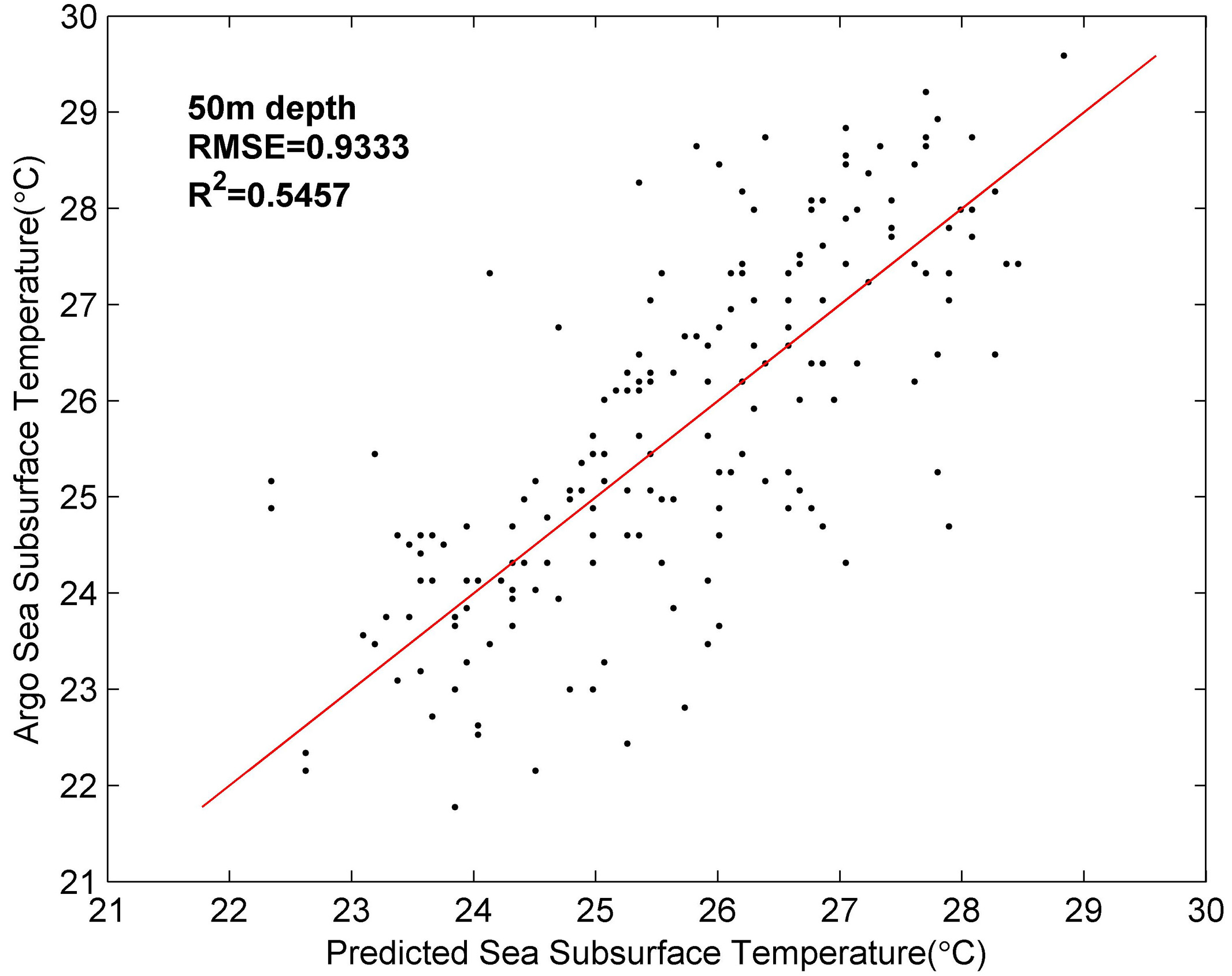}
\end{center}
\vspace*{-6mm}
\caption{Predicted temperature at 50m undersea and corresponding Argo data scatter plot.}
\label{fig_scatter_50m} 
\vspace*{-4mm}
\end{figure}

\begin{figure}[!th]
\vspace*{-2mm}
\begin{center}
\includegraphics[width=0.77\columnwidth,height=0.77\columnwidth]{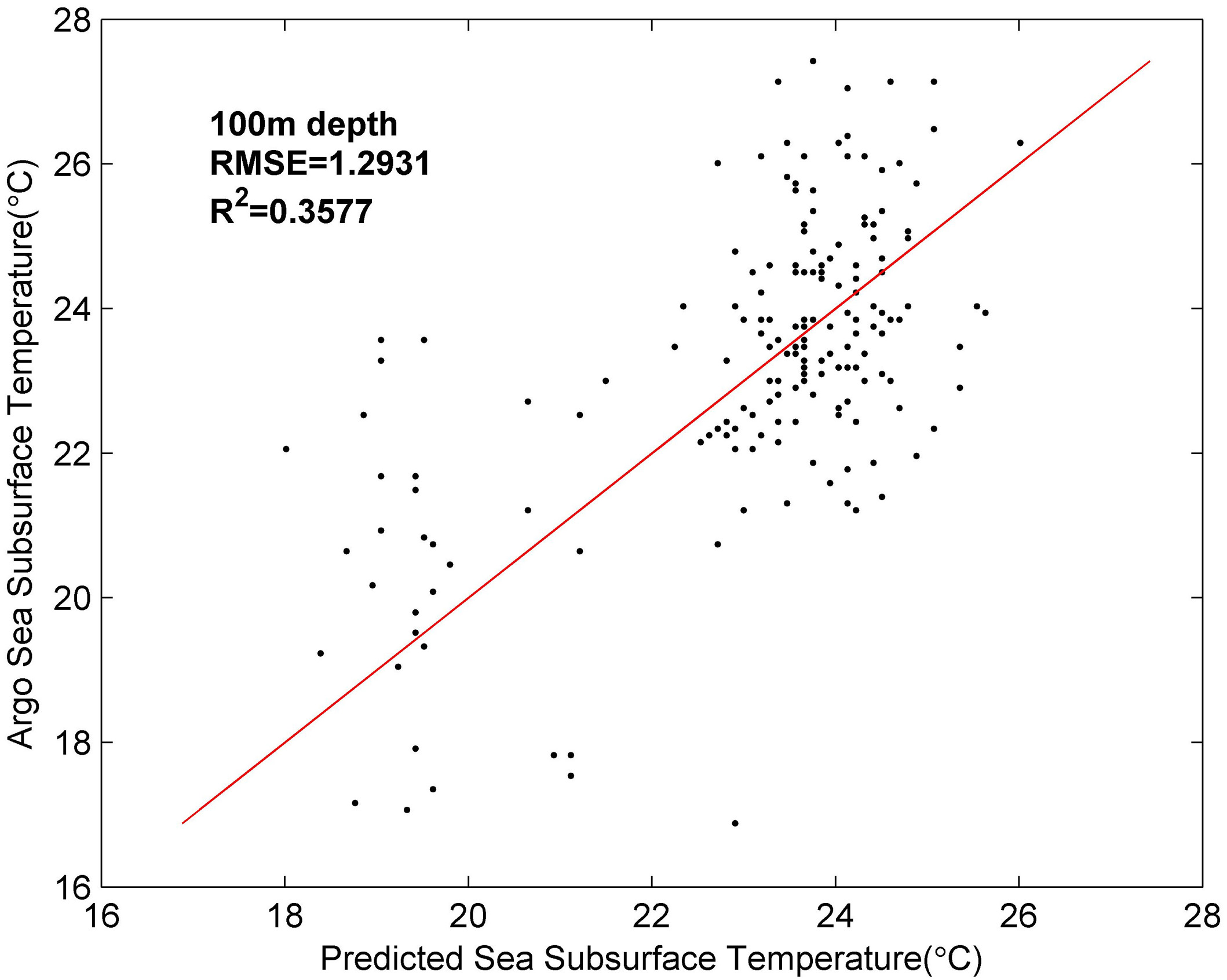}
\end{center}
\vspace*{-6mm}
\caption{Predicted temperature at 100m undersea and corresponding Argo data scatter plot.}
\label{fig_scatter_100m} 
%
\begin{center}
\includegraphics[width=0.77\columnwidth,height=0.77\columnwidth]{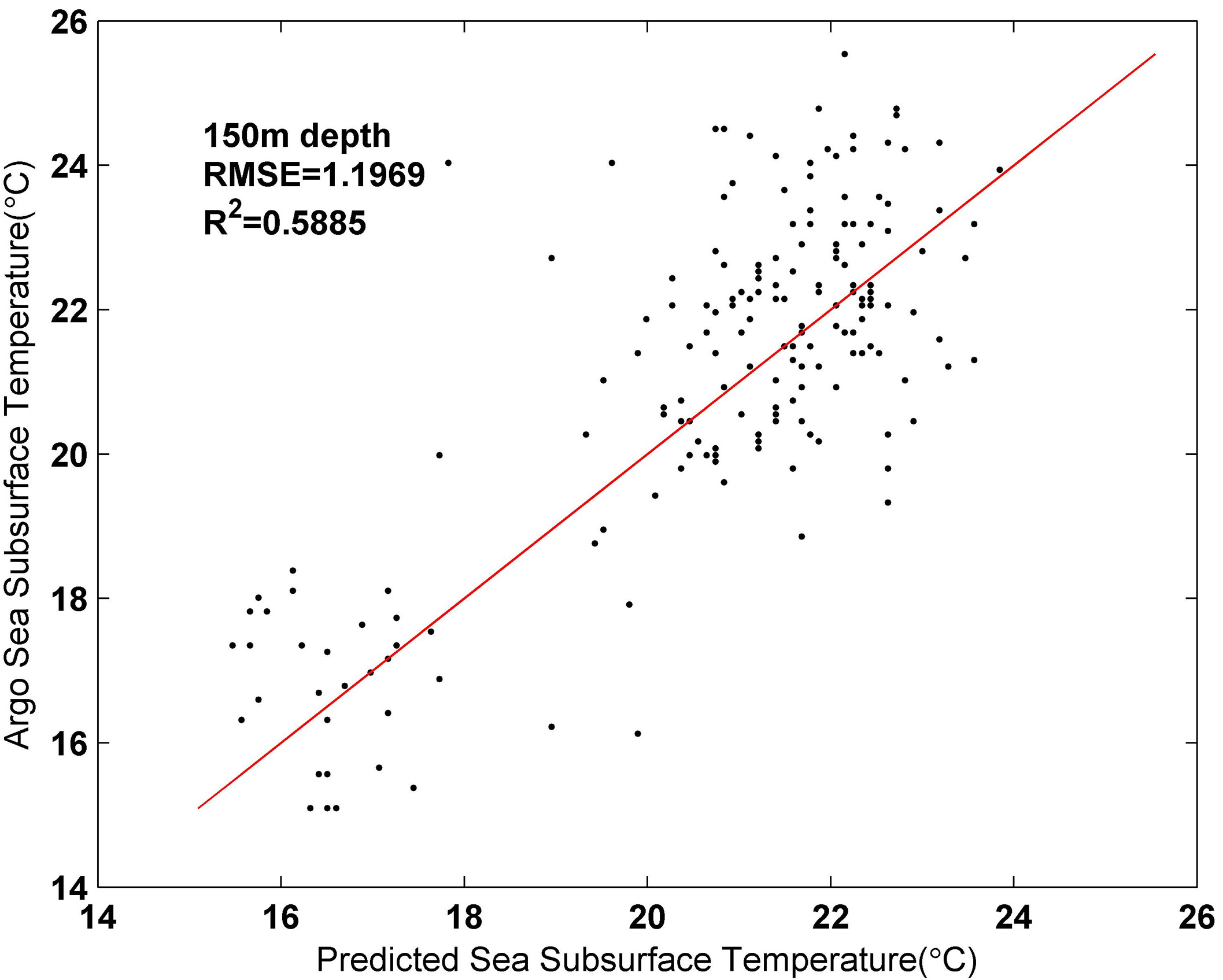}
\end{center}
\vspace*{-6mm}
\caption{Predicted temperature at 150m undersea and corresponding Argo data scatter plot.}
\label{fig_scatter_150m} 
\vspace*{-4mm}
\end{figure}

\subsection{Experimental Results and Analysis}\label{S4.4}

For the Argo data from January 2004 to April 2007, after removing the invalid data, we obtain 180 daily temperature observation values. We compare the predicted results with these 180 remaining Argo observational data. Fig.~\ref{fig_model_argo_50m} compares the predicted temperature at 50m undersea with the corresponding Argo data. It can be observed that the predicted results of the proposed method fit well the Argo data. Similarly, the Argo data and the corresponding predicted temperatures at 100m undersea and 150m undersea are illustrated in Figs.~\ref{fig_model_argo_100m} and \ref{fig_model_argo_150m}, respectively. These results demonstrate that the proposed method can generate reliable and accurate temperature predictions at different depths of the sea.

\begin{figure*}[tp!]
\vspace*{-4mm}
\begin{center}
\includegraphics[width=0.9\linewidth,angle=0]{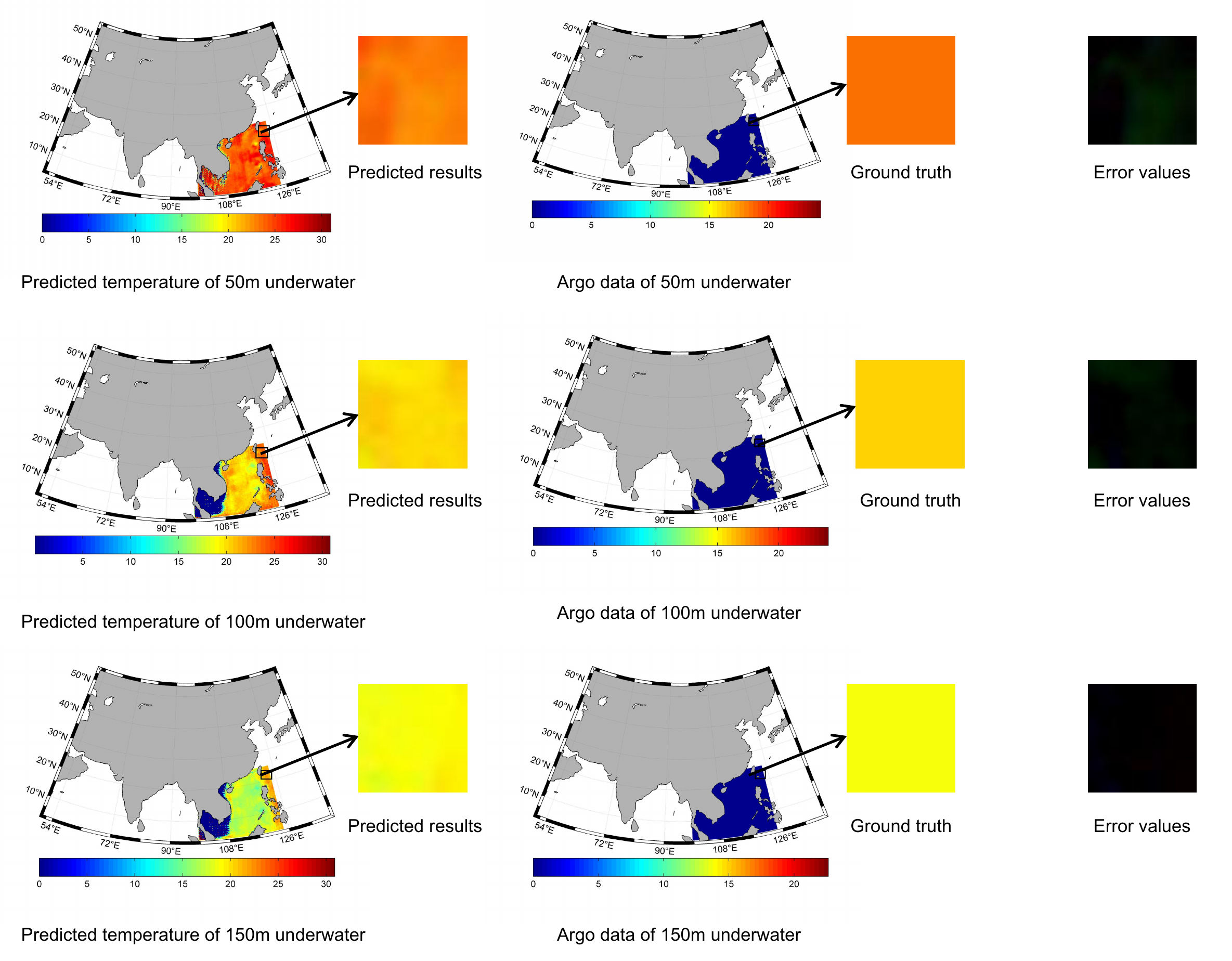}
\end{center}
\vspace*{-4mm}
\caption{Display of predicted temperature values and measurements at different depths on November 9, 2006.}
\label{fig_visual_res} 
\vspace*{-4mm}
\end{figure*}

A correlation scatter plot between the predicted temperature at 50m undersea and the Argo data is depicted in Fig.~\ref{fig_scatter_50m}. If the data points are more evenly and densely distributed near the diagonal red line, the prediction result is better. Similar scatter plots of the prediction results at 100m undersea and 150m undersea are shown in Figs.~\ref{fig_scatter_100m} and \ref{fig_scatter_150m}, respectively. As can be observed, the prediction results at 50m undersea are better than the results at 100m and 150m undersea. Evidently, as depth increases, the prediction accuracy decreases.

Fig.~\ref{fig_visual_res} displays the temperatures predicted by the proposed method at different depths (50m, 100m and 150m) together with the corresponding Argo observations on November 9, 2006. The visual results show that the predicted results by the proposed method are very close to the ground truth Argo data. This demonstrates that the proposed method is reliable and accurate.

The temperature prediction experiment for each model is repeated 10 independent runs with different random initializations. We summarize the temperature prediction results, presented as average$\pm$standard deviation (STD), of different methods in Table~\ref{table_res_all}, where PGpix2pix and PGcycleGAN are the pix2pix method with the numerical model and the CycleGAN method with the numerical model, respectively. Note that applying neural networks, such as pix2pix and CycleGAN, without considering the numerical model is incapable of predicting daily sea subsurface temperature effectively, owing to very limited observational data. Consequently, we have to adopt our idea of physics guided (PG) enhancement by combining neural networks with numerical model. To compare with the methods in \cite{jia_PGRNN} and \cite{zhang19_grsl}, we adopt ConvLSTM model to replace the RNN model for acquiring the sea subsurface temperature prediction in the whole area of China South Sea, which we refer to as PGConvLSTM. Specifically, we train the ConvLSTM model by using the same training mode as ours and removing the mapping from the surface temperature to the subsurface temperature learned by the GAN model as the works \cite{jia_PGRNN,zhang19_grsl} did. In our proposed model, we pre-train the GAN on the numerical model data and then fine-tune the GAN model with the observational data. To compare with this two-stage training, we also simply concatenate the numerical simulation data onto the observational data together to train the GAN, which we refer to as PGsim. Our framework uses the physics loss to automatically encodes the knowledge of ocean physics into the modeling process. In addition to our method with physics loss (Our method with PLoss), the results of our method without physics loss (Our method without PLoss) are also shown in Table~\ref{table_res_all}.

The results of the PGConvLSTM are poor, as this approach does not exploit the mapping from the surface temperature to the subsurface temperature learned by the GAN model \cite{jia_PGRNN,zhang19_grsl}. This demonstrates that this mapping is essential in the prediction of the daily subsurface temperature. Our proposed GAN based framework effectively exploits the merits of both the numerical model and neural network and can learn the map from the surface to the subsurface well through the proposed two-stage training. By contrast, simply concatenating the numerical data and the observational data together to train the model (PGsim) is less accurate than our approach. It can be seen from Table~\ref{table_res_all} that our proposed method with the physics loss attains the best performance. In terms of RMSE, it outperforms PGpix2pix by 0.0126, 0.0407 and 0.056 ($^{\circ}$C) for predicting the sea temperatures 50m, 100m and 150m undersea, respectively. In terms of $R^2$ statistic, our method outperforms PGpix2pix by 0.0163, 0.0376 and 0.0673 for predicting the sea subsurface temperatures at these depths, respectively. Additionally, our method and PGpix2pix have similar STDs for the both performance metrics. Also observe that our method with physics loss outperforms the one without physics loss. Hence, the experimental results clearly demonstrate that the proposed method is capable of enhancing the daily sea subsurface temperature prediction over the existing state-of-the-art methods.

Currently, only the traditional assimilation method can predict the daily sea subsurface temperature. Our proposed method is the first which can significantly improve the accuracy of the daily sea subsurface temperature
prediction compared with the assimilation method. We believe that exploiting the numerical model data and two-stage training mode that we propose are essential to perform the daily sea subsurface temperature prediction task. Multi-task learning is integrated into the proposed method to enable the prediction of the temperatures at 50\,m, 100\,m and 150\,m underwater simultaneously. A physics-based loss is also added to our model to further improve the the accuracy of the daily sea subsurface temperature prediction. The experimental results have verified the effectiveness of our proposed method. Furthermore, our proposed framework also benefits other existing neural network based methods. Although only applying the pix2pix framework or other neural network is incapable of predicting the daily sea subsurface temperature effectively owing to the very limited observational data, by adopting our idea of combining neural network and the numerical model, PGpix2pix becomes capable of significantly outperforming the assimilation method. The experimental results confirm that our proposed method outperforms the existing state-of-the-art methods. Compared with PGpix2pix, although the performance gain is small, our method can predict the temperatures for all the target depths simultaneously, while PGpix2pix needs multiple models to predict the temperatures of different depths.

In our experiments, we have to discard a lot of data since not every daily Argo data is valid. Clearly, by using more usable data sets to provide sufficient training and testing data, the accuracy of prediction can further be improved. In addition, our experimental results will also be enhanced by looking for better quality remote sensed images. 

\section{Conclusions and Future Work}\label{S5}

In this paper, we have proposed a novel GAN-based framework for challenging daily sea subsurface temperature prediction. In our method,  a physics-based numerical model is employed in a GAN to acquire the simplified physical laws at different ocean depths, and observational data are used for fine-tuning the model parameters to obtain better prediction results. Our method has effectively exploited the complementary merit of physics-based numerical model and observational data based neural network. Moreover, a physics-based loss based on a mask has been employed, which leads to improved prediction performance. The experimental results have demonstrated that the proposed method can achieve better performance in daily sea subsurface temperature prediction compared with the state-of-the-art baselines.

In the future, we plan to extend our work to temporal dimension with better quality and large scales traits, which will provide more information to further improve the prediction accuracy. In addition, we also plan to investigate the use of several self-attention networks to enhance the overall performance of our model.

\end{document}